\documentclass[review]{elsarticle}

\usepackage{lineno,hyperref}
\usepackage{comment}
\usepackage{amsmath}
\usepackage{graphicx}
\usepackage{multirow}
\usepackage{subfigure}
\usepackage{tabularx}
\usepackage{booktabs}

\modulolinenumbers[5]
\usepackage{color,xcolor}

\journal{Journal of \LaTeX\ Templates}

\begin{document}

\begin{frontmatter}

\title{Continual Learning via Inter-Task Synaptic Mapping}
\tnotetext[mytitlenote]{Corresponding Author}
\tnotetext[mytitlenote1]{Equal Contribution}
\author[mymainaddress,mysecondaryaddress]{Mao Fubing$^{**}$}

\author[mysecondaryaddress]{Weng Weiwei$^{**}$}
\author[mysecondaryaddress]{Mahardhika Pratama*$^{**}$}
\author[mythirdaddress]{Edward Yapp Kien Yee}

\address[mymainaddress]{National Engineering Research Center for Big Data Technology and System, Services Computing Technology and System Lab, Cluster and Grid Computing Lab, School of Computer Science and Technology, Huazhong University of Science and Technology, Wuhan 430074,China}
\address[mysecondaryaddress]{School of Computer Science and Engineering, Nanyang Technological University, Singapore}
\address[mythirdaddress]{Singapore Institute of Manufacturing Technology, A*Star, Singapore}

\begin{abstract}
Learning from streaming tasks leads a model to catastrophically erase unique experiences it absorbs from previous episodes. While regularization techniques such as LWF, SI, EWC have proven themselves as an effective avenue to overcome this issue by constraining important parameters of old tasks from changing when accepting new concepts, these approaches do not exploit common information of each task which can be shared to existing neurons. As a result, they do not scale well to large-scale problems since the parameter importance variables quickly explode. An Inter-Task Synaptic Mapping (ISYANA) is proposed here to underpin knowledge retention for continual learning. ISYANA combines task-to-neuron relationship as well as concept-to-concept relationship such that it prevents a neuron to embrace distinct concepts while merely accepting relevant concept. Numerical study in the benchmark continual learning problems has been carried out followed by comparison against prominent continual learning algorithms. ISYANA exhibits competitive performance compared to state of the arts. Codes of ISYANA is made available in https://github.com/ContinualAL/ISYANAKBS.       
\end{abstract}

\begin{keyword}
continual learning, lifelong learning, catastrophic forgetting
\end{keyword}

\end{frontmatter}


\section{Introduction}

Continual learning aims to emulate the underlying trait of natural learning to learn various tasks on the fly without losing competence for what has been achieved in the past~\cite{vandeven2019three,vandeven2018generative,gem2017paz}. Ideally, already owned skill should ease the learning process of different-but-related tasks in the future. This problem appears as an extension of data stream learning~\cite{learning2010peipei} where a learner must not only demonstrate adaptive and evolving aptitudes to handle non-stationary environments but also possess the knowledge retention property such that it gains more intelligence as the increase of tasks being learned~\cite{overcome2017lee}. 

The underlying challenge of continual learning lies in the catastrophic forgetting problem where learning a new task catastrophically replaces old knowledge with new one thereby losing its relevance to handle old tasks. The regularization approach~\cite{overcome2017lee,kirkpatrick2016overcoming} is among several well-known methods to resolve the catastrophic forgetting problem of the continual learning ~\cite{overcome2017lee,kirkpatrick2016overcoming}. The key idea is to prevent important parameters from being perturbed by learning a new task.  Elastic Weight Consolidation (EWC)~\cite{kirkpatrick2016overcoming} is one pioneering work in this area where the unique trait is seen in the application of Fisher information matrix to construct the parameter importance matrix while integrating the L-2 norm like regularization approach~\cite{overcome2017lee,kirkpatrick2016overcoming}. The synaptic intelligence (SI) method is proposed to address expensive computation of parameter importance matrix via the Fisher information matrix~\cite{zenke2017continual}. The accumulation of loss for every training sample is used to inform the significance of each synaptic. This work has been extended in \cite{schwarz2018progress} called onlineEWC using the Laplace approximation. The learning without forgetting (LWF) approach is slightly different where it uses a joint optimization procedure between the cross entropy loss of current task and the knowledge distillation loss~\cite{li2016learning}. It is later found that the knowledge distillation loss can be replaced by the cross-entropy loss~\cite{li2016learning}. Memory Aware Synapses (MAS) is a regularization-based approach where the importance of network parameters are calculated in an unsupervised and online manner. Another prominent work is proposed in \cite{neuron_level} where the regularization mechanism is performed in the neuron level instead of the synaptic level by measuring neuron importance and governing learning rates.

The main limitation of these approaches lies in the absence of inter-task relationship where a new task might share some commonalities to previous tasks. That is, a neuron might still accept new information or even contribute to understand the current task. Since a new task is embraced by those which do not contribute too much to the previous tasks, one should guarantee enough network capacity to handle all tasks otherwise new synapses have to be grown. Another challenge also exists in scaling the parameter importance matrix which can quickly explode in the large-scale problems thus causing the unlearning effect since it has to be accumulated across all tasks~\cite{parisi2018continual} unless an independent parameter importance matrix has to be created from scratch for every task.  

An Inter-Task-Synaptic-Mapping (ISYANA) is proposed here to address the catastrophic forgetting problem of the continual learning\frenchspacing. ISYANA is built upon the task-to-synapses and task-to-task mappers. The task-to-synapses module provides a relational mapping between each hidden node and all tasks determining the acceptable type of information it can absorb. The task-to-task component exhibits the inter-task relationship thus revealing the complete problem structure and the commonalities of each task. This module is capable of providing a sort of neighborhood degree of a sample to all tasks which can be linked to the task-to-synapses module. The relationship of a sample to previous tasks is assessed by checking its proximity to previous tasks. A node is frozen by reducing its learning rate provided that it has low relevance to the current task and the current task characterizes low mutual information to the old tasks. In a nutshell, the parameter importance matrix is crafted from the combination of the task-to-task and task-to-synapses relationships. The catastrophic forgetting is handled in the neuron level due to the fact where a connective weight plays little role to a learning problem unless it is combined with other weights to construct a neuron,. i.e., hierarchical structure of a network structure.  

The concept-to-concept or task-to-task module is devised by the cluster guided mechanism in the deep latent space. It puts forward a meta-network crafted from an extension of the deep clustering network~\cite{yang2016kmeansfriendly,Guo2017DeepCW}. That is, a class-specific cluster is designed in the deep latent space. Note that the clustering approach is carried out fully in the per-class manners such that learning a new task should not disturb the representation of old tasks. The concept-to-concept module applies \textbf{the growing and consolidation phases}. The growing phase constructs a set of clusters describing the current task based on the class label of each task. The consolidation phase is undertaken after the growing phase by calculating the mean values of the clusters to bound the network complexity where the mean values of the clusters exhibit a complete summary of already seen data points. In other words, there does not exist \textbf{any requirement to store representation layer of each task} which does not sustain for large-scale problems. The inter-task similarity is calculated via the KL divergence approach among all clusters between tasks. 

The task-to-synapses module complements the task-to-task module where it functions as the one-to-one mapping between a neuron and a target class.  It measures the mutual information of a node to each task making possible for a neuron to still accept the current concept notwithstanding that it is highly relevant to previously seen tasks. In short, a node can be shared by different but related tasks. All of which can be done in the online fashion without being prone to \textbf{the issue of exploding parameter importance matrix} thereby improving its feasibility for a high number of tasks. The advantage of ISYANA has been numerically validated in benchmark continual learning problems. It is compared with the prominent continual learning algorithms where it demonstrates competitive performance compared to recently published algorithms. Codes of ISYANA is made available in https://github.com/ContinualAL/ISYANAKBS.

\section{Related Works}

ISYANA is derived from recent works on regularization principle for continual learning where the important parameters of old tasks are precluded from accepting the new concept to overcome the catastrophic forgetting problem. In a nutshell, the regularization-based approach works in the synaptic level formulated as a joint optimization problem as follows:
\begin{equation}\label{parameter}
    L=L(Y,\hat{Y})+\frac{\alpha}{2}Z(\theta-\theta^{*})^2
\end{equation}
where $\alpha$ is a regularization constant while $\theta,\theta^{*}$ are the existing network parameters and the previous optimal parameters trained at $T-1$ task. That is, all network parameters are embedded to $\theta$. The key component lies in the parameter importance matrix $Z$ measuring the contribution of each parameter in the previous task. $Z$ is computed from the Fisher information matrix concept in EWC while the synaptic intelligence (SI) utilizes the combination of network gradient and parameter movement to deduce the network significance. Nonetheless, the main bottleneck of this approach is its application to the large-scale problems since the parameter importance matrix $Z_t$ has to be calculated for each task and combined to previous task $Z_{t-1}+Z_{t}$ when being applied to the $t+1$ task~\cite{kirkpatrick2016overcoming,zenke2017continual}. This issue causes the explosion of $Z$ in long run thereby leading to the unlearning effect of old parameters and hindering for \textbf{the positive forward and backward transfer mechanism}. It is almost impossible to be resolved without the introduction of new parameters to handle a new task. The practical solution is with the normalization technique~\cite{overcome2017lee,kirkpatrick2016overcoming}. Another solution is by purposely insert the forgetting factor $\gamma_t Z_{t-1}+Z_{t}$~\cite{overcome2017lee,kirkpatrick2016overcoming}. 

Our approach, ISYANA, offers an alternative solution where the inter-task relationship is taken into account while considering the relevance between synaptic and a task. The goal is to allow a node to learn multiple relevant tasks thereby underpinning the positive forward and backward transfer. That is, the two mappings reveal the common feature of a task which can be shared across  similar tasks while retaining private information of a particular task. Furthermore, the inter-task mapping also eliminates the accumulation of parameter importance mapping thus coping with the unlearning effect with the absence of controlled forgetting mechanism~\cite{overcome2017lee,kirkpatrick2016overcoming}. 

\section{ISYANA: Inter-Task Synaptic Mapping}
\label{sec:intertaskSynapticMap}
\subsection{Problem Formulation}
\noindent\textbf{Formal Definition}: Continual learning aims to build a predictive model which can handle the never-ending arrival of tasks. Consider $T_t=(X_t,Y_t)\in\Re^{N}$ is the $t-th$ task consisting of a pair of data sample $X_n\in\Re^{u},Y_n\in\Re^{m}$ as input and target data points where $u,N,m$ are the dimension of input space, data space and the number of classes seen thus far respectively. The number of tasks, $T$, is unknown in practise. The typical characteristic of continual learning calls for a learner with light computational and space complexities which must not be a factor of the number of tasks and data samples. That is, a task cannot be revisited again in the future once learned.

The continual learning problem suffers from non-stationary environments where a task is not drawn from static concept. That is, there exists the issue of drift where the concept of $t-th$ task is drifted toward a new distribution in the next task $P(Y|X)_t\neq P(Y|X)_{t+1}$~\cite{pratama2019automatic}. Another common issue lies in the incremental class problem where a new task $T_t$ is presented with a set of new classes $m'$ while being completely isolated from the old classes $m$ previously seen in the old tasks. In other words, $m'$ does not appear together with the previous classes. This case leads to the catastrophic forgetting problem in which a model loses its aptitude in dealing with previously seen tasks. 

The typical characteristic of the continual learning problem distinguishing itself from other learning problems is seen in the knowledge retention requirement. That is, a continual learner must be prepared to be queried by any samples following the concepts or classes of old tasks meaning that learning a new task must not catastrophically erase its past knowledge base~\cite{overcome2017lee,kirkpatrick2016overcoming}. The key challenge is to find tradeoff points addressing all tasks without seeing them in the same basket. A regularization approach is chosen here to cope with the catastrophic forgetting problem since it incurs low complexity being independent to the problem size - no old samples or representation layers of each task have to be stored.

\noindent\textbf{Definition of Network Structure}: ISYANA is integrated in the context of multi-layer perceptron (MLP) network where $n_l,L$ stand for the number of hidden nodes of the $l-th$ layer and the total number of layers respectively. A $l-th$ layer is defined as $h_{l}=s(W_{in}^ls_{i,(l-1)}+b_{l})$ where its node is assigned as the sigmoid function. $W_{in}^l\in\Re^{n_l\times n_{l-1}},b_l\in\Re^{n_l}$ are the connective weights and bias. The softmax function $\xi=\frac{\exp{(o_j)}}{\sum_{j=1}^{m}(o_j)}\in\Re^{m}$ is applied at the last layer $L$ where $o=W_{out}h_L+c$. $W_{out}\in\Re^{m\times n_L},c\in\Re^{m}$ denote the output weight and the output bias respectively. 
\subsection{Algorithm}
ISYANA offers an efficient alternative for construction of the parameter importance matrix $Z$ where the importance of a neuron is determined from its mutual information to the current task as well as to the old tasks thereby supporting the positive forward and backward transfer provided its high mutual information. ISYANA is underpinned by the task-to-task mapping and the synaptic-to-task mapping in deriving the parameter importance matrix. The importance of the $j-th$ node of the $l-th$ layer in respect to a current task is formalized as the linear combination of its task-to-task and synaptic-to-task relationships as follows:
\begin{equation}\label{significance}
    \phi_{j,l}^{t}=TT_{t,1}*ST_{j,l}^{1}+TT_{t,2}*ST_{j,l}^{2}+...+TT_{t,m}*ST_{j,l}^{m}
\end{equation}
where $TT_{t,o}\in[0,1]$ represents the task-to-task mapping of the current $t-th$ task to $o$-th class while $ST_{j,l}^{o}\in [0,1]$ defines the synaptic-to-task mapping of the $j-th$ node of $l-th$ layer to the current $o-th$ class. $TT_{t,m_t}=1$ for any $m_t$ classes associated to the current $t-th$ task. That is, the relationship of the current task to itself is maximum. In our implementation, the $t-th$ task is formulated by its target classes following the normal distribution. (\ref{significance}) is bounded in $[0,m]$ making the scaling convenient to be undertaken. The parameter importance matrix $Z_t$ for the $t-th$ task is defined as follows:
\begin{equation}\label{Z}
    Z_t=
    \begin{bmatrix}
    \exp{(-\phi_{1,1}^{t})} & ... & \exp{(-\phi_{n_1,1}^{t})}\\
      \exp{(-\phi_{1,2}^{t})} & ... & \exp{(-\phi_{n_2,2}^{t})}\\
      ......\\
            \exp{(-\phi_{1,L}^{t})} & ... & \exp{(-\phi_{n_L,L}^{t})}\\
\end{bmatrix}
\end{equation}
The parameter importance matrix $Z$ can be conveniently integrated to the learning rate with respect to the loss gradient. In other words, the catastrophic forgetting is controlled in the neuron level rather than in the synaptic level as per (\ref{parameter}) because of the hierarchical nature of the deep neural network \cite{neuron_level}. Instead of fixed learning rate, a node-varying learning rate is introduced as follows:
\begin{equation}\label{ALR}
\begin{aligned}
  \eta = a*\exp{(-b*Z+c)}
\end{aligned}
\end{equation}
where a, b and c are constant parameter that helps to adjust learning rate. The node-varying learning rate is determined by the parameter importance matrix $Z$ accordingly. The exponential term is incorporated into (\ref{Z}) since the regularization magnitude should be inversely proportional to the importance of a hidden node. The higher the importance of a node the lower the loss gradient is induced. A node accepts the current concept and enables positive forward and backward transfer if it receives relevant information. That is, its importance to the current task increases if the current task shows common information to classes in which a node is relevant to. On the other hand, its importance diminishes in the case of unique task to which a node is associated to. This strategy reflects to the random initialization property of deep neural network making possible for a node to converge to particular classes. This concept follows the neuron-level plasticity control in \cite{neuron_level} but our approach not only takes into account the parameter importance but also the inter-task relationship.  

\begin{figure}[!ht]
\centering
\includegraphics[width=1.0\columnwidth]{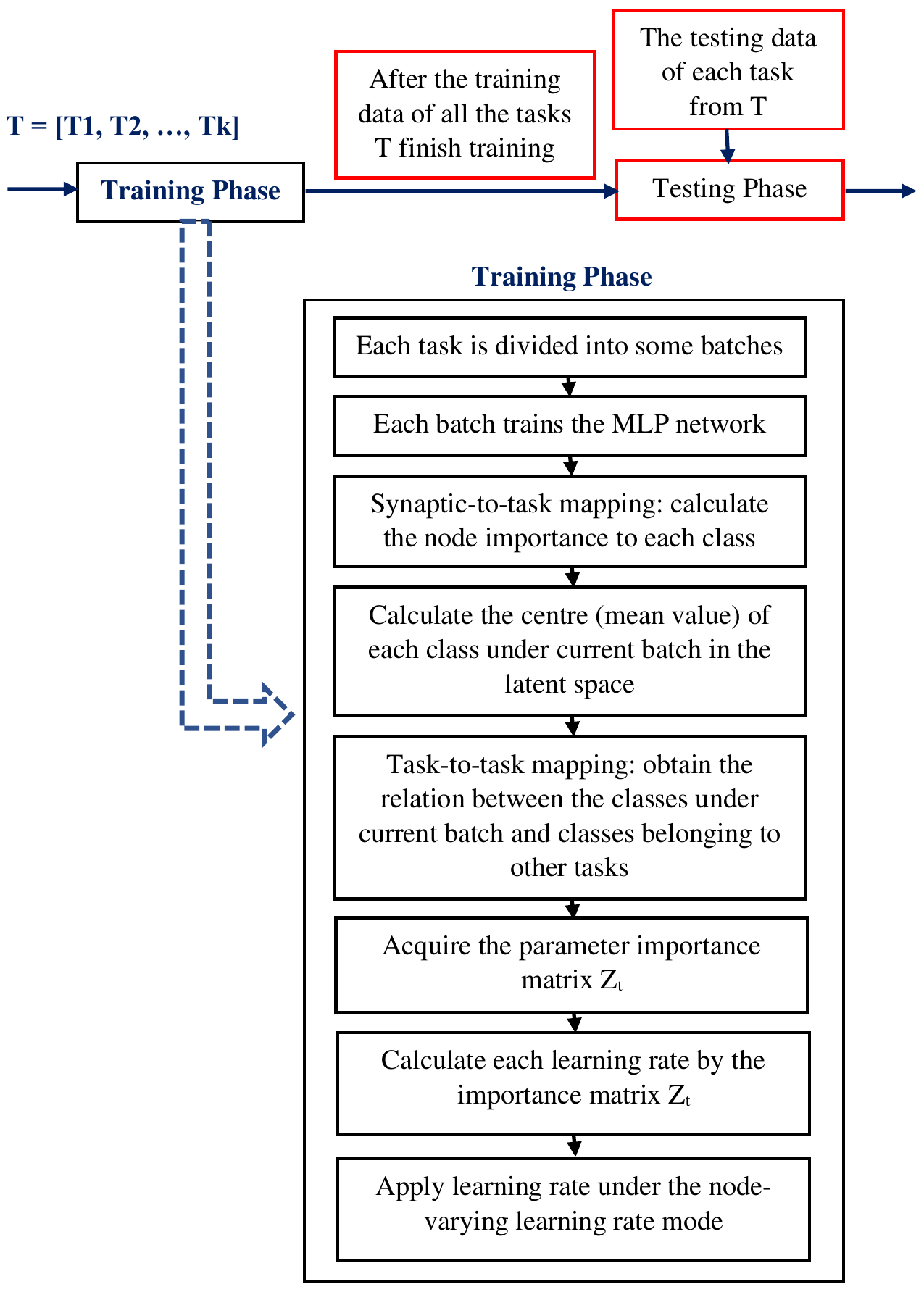}
\caption{The overall flow framework of ISYANA. The training phase contains several steps.}
\label{fig:isyanaflow}
\end{figure}

Fig.~\ref{fig:isyanaflow} shows the overall learning policy of our approach ISYANA, consisting of training phase and testing phase. There is a set of tasks $T = {T_{1},T_{2}, ...,T_{k}}$. Each task contains several classes and each task trains the network in sequence. When the training process of all the tasks is finished, we feed the testing data of each task to test the final model and evaluate them according to three criterion: the average accuracy, the backward transfer and the forward transfer.

Training Phase of Fig.~\ref{fig:isyanaflow} introduces the steps of the training phase. Firstly, each task is divided into some mini batches used to train the MLP network. Secondly, we calculate the node importance to each class (current output class). This part belongs to synaptic-to-task mapping described in Section~\ref{sec:intertaskSynapticMap}. Thirdly, we adopt a stacked autoencoder to map each batch to the latent space. We calculate the center for each target class under the current batch in the latent space. Then, we obtain the relation between the classes belonging to the current batch and classes belongs to other tasks. This component is attributed to task-to-task mapping introduced in Section~\ref{sec:intertaskSynapticMap}. Fourthly, we acquire the parameter importance matrix $Z_t$ mentioned above which will be incorporated in (\ref{ALR}) in the learning rate of stochastic gradient descent (SGD) method. 

\subsection{Synaptic-to-Task Mapping}
This component examines the relevance of a node to target classes thus enabling the positive forward and backward transfer mechanism~\cite{gem2017paz}. That is, a node can still learn the current task despite being important to the old task if it exhibits strong mutual information. It is inspired by the existence of common information of different tasks where learning old task might expedite the learning process of a new task. The relevance of the $j-th$ node of $l-th$ layer to $o-th$ target class is formulated by applying the symmetrical uncertainty measure~\cite{OENTARYO201112066} as follows:
\begin{equation}
    ST_{j,l}^{o}=\frac{2I(h_{j,l},o)}{H(h_{j,l})+H(o)}
\end{equation}
where $I(h_{j,l},o)$ denotes the information gain or the mutual information of the hidden node $h_{j,l}$ and the $o-th$ target class (network output) while $H(h_{j,l}),H(o)$ respectively stand for the entropy of the $j-th$ node of $l-th$ layer and the $o-th$ target class (network output). The symmetrical uncertainty measure is chosen here because it is simple and has low bias for multi-valued features~\cite{OENTARYO201112066}. Calculation of information gain and entropy are typically expensive. By assuming normal distribution, the notion of differential entropy is adopted here where the information gain and entropy are respectively defined as $I(x,y)=-\frac{1}{2}log(1-\rho(x,y)^{2}),H(x)=\frac{1}{2}(1+log(2\pi\sigma_{x}^2))$~\cite{OENTARYO201112066}. $\rho(x,y),\sigma_{x}^{2}$ refer to the Pearson's correlation measure between two variables and the variance of a random variable $x$. Note that both of them can be calculated recursively in the incremental fashion. 

The synaptic-to-task mapping is not only applicable in the incremental class setting but also effective in the case of the concept drift albeit no introduction of new classes. In particular, a node supporting the $o-th$ class is not triggered due to low learning rate $P(Y|X)_t\neq P(Y|X)_{t+1}$ if the new concept has little relationship to the previous class - the distributional change leads to the change of classification boundary. The concept change diminishes the relevance of a node if its relevance to another class is not substantiated. This mechanism only needs to store the synapses and bias of deep neural networks $[W_{in},b],[W_{out},c]$.
\subsection{Task-to-Task Mapping}
This module is meant to learn the commonalities across tasks to explore the possibilities of positive forward transfer if a current task shares similarities to the previously seen tasks. The compatibility of the current task to the old tasks is examined by measuring the divergence of their class distributions. A sample of the current task having strong relationship to the old tasks can be shared to the existing nodes playing important roles to the old tasks. In other words, a node do not stop its learning process although it is found to be pivotal for the previous tasks. Notwithstanding that the synaptic-to-task mapping already analyzes the relevance of each node to all tasks, the task-to-task mapping is still required to confirm a learning process of a new task. Furthermore, the task-to-task mapping is relatively stable compared to the synaptic-to-task mapping because its calculation is not affected by other tasks. Once a representation of a class is created, it is frozen. It is only utilized to execute the task-to-task mapping.  

It is built upon a flexible deep clustering method summarizing a task into a number of clusters in the latent space. It is inspired by the idea of deep clustering network in~\cite{yang2016kmeansfriendly,Guo2017DeepCW} where the latent representation is constructed from the stacked auto encoder~\cite{yang2016kmeansfriendly,Guo2017DeepCW} rather than the simple linear mapping often being trapped in the trivial solution. Our underlying contribution here is to perform clustering according to the labels of the tasks and the total number of clusters are equal to the number of classes among all the tasks. That is, every task is formulated as a cluster of normal distribution thereby enabling to evaluate their commonalities. 

\noindent\textbf{Construction of Cluster-Friendly Latent Space}: the latent space is learned under the framework of stacked auto-encoder with $L_{ae}$ hidden layers. We adopt the same notion of deep clustering network~\cite{yang2016kmeansfriendly,Guo2017DeepCW} where the overall loss function of the clustering network encompasses the reconstruction error as follows:
\begin{equation}\label{clustloss}
    L=L(\hat{x},x)
\end{equation}
(\ref{clustloss}) compresses the input space in the nonlinear fashion via the use of deep autoencoder thereby inducing the cluster-friendly latent space. Furthermore, the application of stacked autoencoder here prevents the trivial solution often encountered by the linear transformation~\cite{yang2016kmeansfriendly,Guo2017DeepCW}. The end-to-end training strategy is carried out in respect to (\ref{clustloss}) for each layer where $[W_{en}^{l_{ae}},b_{en}^{l_ae}]$ stand for the weight and bias of the $l_{ae}-th$ encoder while $[W_{de}^{l_{ae}},b_{de}^{l_{ae}}]$ denote the weight and bias of the $l_{ae}-th$ decoder. Note that the weight of decoder is the inverse mapping of encoder $W_{de}^{l_{ae}}=W_{en}^{l_{ae}^{T}}$. Note that the AE here is trained continuously across all tasks to establish the representation of each class. Furthermore, this mechanism only needs to store the weight and bias of autoencoder (AE) $[W_{de},b_{de}],[W_{en},b_{en}]$. 

\noindent\textbf{Calculation of The Inter-Task Mapping}: once representing a class as $N(\mu,\sigma)$, the relationship between tasks and the target class in the latent space $s_{L_{ae}}$ is formally defined using the Kullback-Leibler divergence measure as follows:
\begin{equation}\label{TT}
\begin{aligned}
    TT_{t,o}= 1/((KL(\mu_{i,1}, \mu_{t,o}) + KL(\mu_{i,2}, \mu_{t,o})\\ +...+KL(\mu_{i,CO}, \mu_{t,o}))/CO)
\end{aligned}
\end{equation}
where KL is Kullback-Leibler Divergence~\cite{Sankaran2016KullbackLeiblerDA} to evaluate the difference between two probability distributions over the same variable $x$~\cite{Sankaran2016KullbackLeiblerDA}. In our work, we assume that each class is represented as the normal distribution. $CO$ denotes the number of clusters belonging to a specific task $i$ evaluated to the $o-th$ target class belonging to a specific task $t$. The higher the value of (\ref{TT}) leads to the higher the relationship of the current task to the old tasks. That is, a sample possibly offers the positive forward transfer mechanism thus being able to be accepted for the current model update - a weak regularization effect is returned. No forgetting case occurs here since the representation of each class is fixed once created. Furthermore, (\ref{TT}) does not depend on the AE representation updated continuously across all tasks. This mechanism only needs to store the mean of each class $\mu_{t,o}$.

\section{Proof of Concept}
This section discusses the experimental procedure of our algorithm in four problems: splitMNIST~\cite{gem2017paz,goodfellow2013empirical}, permuttedMNIST~\cite{gem2017paz}, rotatedMNIST~\cite{jaderberg2015spatial} and omniglot ~\cite{schwarz2018progress}. ISYANA is compared against eight recently published algorithms: Context-dependent Gating (XDG)~\cite{MasseE10467}, Elastic Weight Consolidation (EWC)~\cite{kirkpatrick2016overcoming}, onlineEWC~\cite{gem2017paz,schwarz2018progress}, SI~\cite{zenke2017continual}, LWF~\cite{li2016learning}, Deep Generative Replay (DGR)~\cite{shin2017continual}, Deep Generative Replay distillation \cite{vandeven2019three,shin2017continual} and Averaged Gradient Episodic Memory (A-GEM) \cite{Chaudy2019AGEM}.
\newline\textbf{Simulation Protocol}: the standard evaluation protocol of continual learning~\cite{overcome2017lee,kirkpatrick2016overcoming} is followed here where three evaluation metrics, namely average accuracy, positive forward transfer and positive backward transfer, are applied. It not only evaluates the ISYANA's aptitude in dealing with catastrophic forgetting, but also assesses whether the past knowledge improves the network performance in learning new tasks or the newly observed knowledge in the current task undermines its capability in coping with the previous tasks. Suppose that $S\in\Re^{T\times T}$ stands for an evaluation matrix where $S_{i,j}$ denotes the test accuracy on the task $T_j$ after observing completely the task $T_i$ while $\hat{b}$ is the accuracy vector for each task. The three evaluation metrics~\cite{vandeven2019three,vandeven2018generative,gem2017paz} are mathematically defined as follows:
\begin{equation}\label{acc}
    ACC=\frac{1}{T}\sum_{i=1}^{T}S_{T,i}
\end{equation}
\begin{equation}\label{BWT}
    BWT=\frac{1}{T-1}\sum_{i=1}^{T-1}(S_{T,i}-S_{i,i})
\end{equation}
\begin{equation}\label{FWT}
    FWT=\frac{1}{T-1}\sum_{i=2}^{T}(S_{i-1,i}-\hat{b}_i)
\end{equation}
where $ACC,BWT,FWT$ stand for the average accuracy, the backward transfer and the forward transfer respectively. The $BWT$ and $FWT$ complement $ACC$ if the accuracy returns the same results.
\newline\textbf{Baseline}: the characteristics of eight baseline algorithms are outlined as follows:
\newline\textit{XDG}~\cite{MasseE10467} uses a randomly generated number of hidden nodes to accept new tasks while leaving the remainder of nodes unchanged. 
\newline\textit{EWC}~\cite{kirkpatrick2016overcoming} is a regularization approach for handling the catastrophic forgetting problem where the parameter importance matrix $Z$ is derived from the Fisher information matrix.  
\newline\textit{OnlineEWC}~\cite{schwarz2018progress} is an online variant of EWC where it utilizes the Laplace approximation rather than the posterior approximation. It retains only the latest running sum on the Fisher's information matrix. 
\newline\textit{SI}~\cite{zenke2017continual} is akin to EWC but offers an alternative avenue in computing the parameter importance matrix $Z$. That is, it is calculated using the accumulated gradient of network parameters to reduce the computational burden. 
\newline\textit{LWF}~\cite{li2016learning} prevents the catastrophic forgetting by formulating the joint optimization problem between the cross entropy loss of current task and the knowledge distillation loss.
\newline\textit{DGR}~\cite{shin2017continual} utilizes the information replay mechanism to cope with the catastrophic forgetting problem. It makes use of the generative adversarial network (GAN) principle where there exists a generator and solver for each task. 
\newline\textit{DGR+distill}~\cite{vandeven2019three,shin2017continual} presents an extension of DGR with the knowledge distillation concept. It reduces the computational issue on DGR with the integration of generative model into the main model through the backward connection. 
\newline\textit{A-GEM}~\cite{Chaudy2019AGEM} is an extension of Generative Episodic Memory \cite{gem2017paz} where it applies some modification of loss function of GEM. 
\newline All algorithms are configured under the same network structure and utilizes the implementation~\cite{vandeven2019three,vandeven2018generative}. In a nutshell, the hyperparameters of consolidated algorithms are exhibited in Table \ref{hyper-para mnist} and \ref{Hyper-omniglot}. Additional details of the experiments can be found in the Appendix.  
\newline\textbf{Datasets}: the characteristics of the four baseline problems are detailed in the following. 
\newline\textit{Rotated MNIST} is an extension of the original MNIST problem~\cite{gem2017paz,goodfellow2013empirical,jaderberg2015spatial} having non-stationary traits. The changing characteristics are induced by dynamic rotations between $[0,180]$ degrees. 
\newline\textit{Permutted MNIST} presents the non-stationary version of MNIST problem by applying the fixed pixel permutation. Both permutation and rotation of the permuttedMNIST and the rotatedMNIST are unrelated for each task and represent the concept drift while having fixed target classes across tasks. 
\newline\textit{Split MNIST} features the incremental class problem where the full MNIST problem is divided into 5 subsets or tasks of disjoint digits: $T_1\in[0,1],T_2\in[2,3], T_3\in[4,5], T_4\in[6,7], T_5\in[8,9]$. \newline\textit{Omniglot} is a popular benchmark problem of the few-shot learning modified for the continual learning. It presents the continual learning of handwritten characters of 50 alphabets \cite{schwarz2018progress}. Each alphabet is considered as a separate task. Because of its size having 500 classes in total, this problem is capable of testing the scalability of continual learner. 
 
\begin{figure}[t]
\centering \subfigure[Average Accuracy.]{
   \label{fig:splitacc}
   \includegraphics[width=1.\columnwidth]{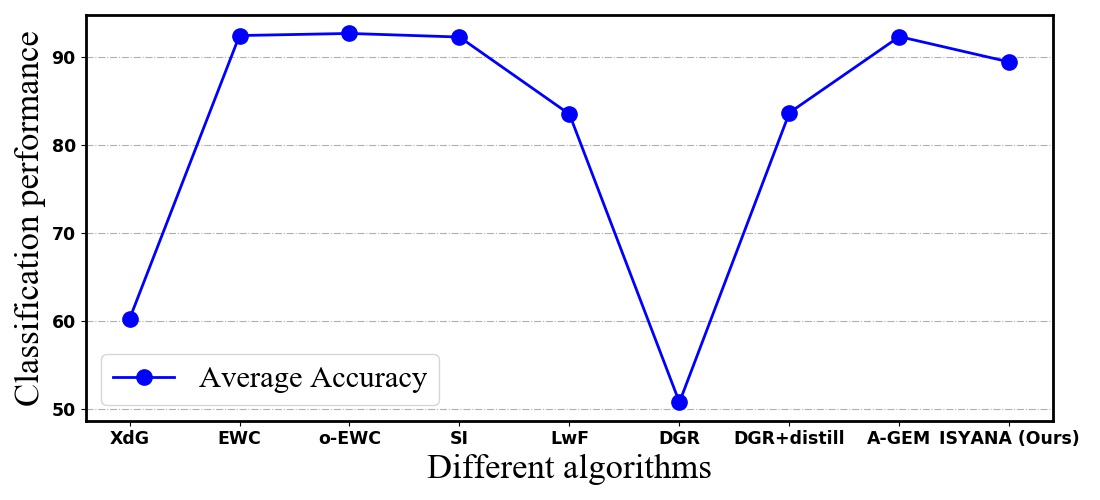}
} \subfigure[Forward Transfer.]{
   \label{fig:splitfwt}
   \includegraphics[width=1.0\columnwidth]{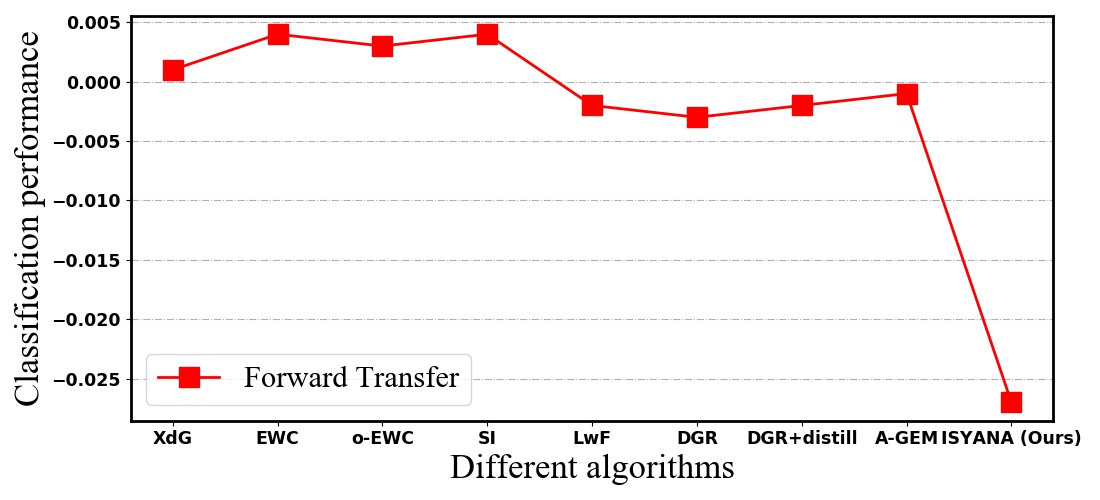}
} \subfigure[Backward Transfer.]{
   \label{fig:splitbwt}
   \includegraphics[width=1.0\columnwidth]{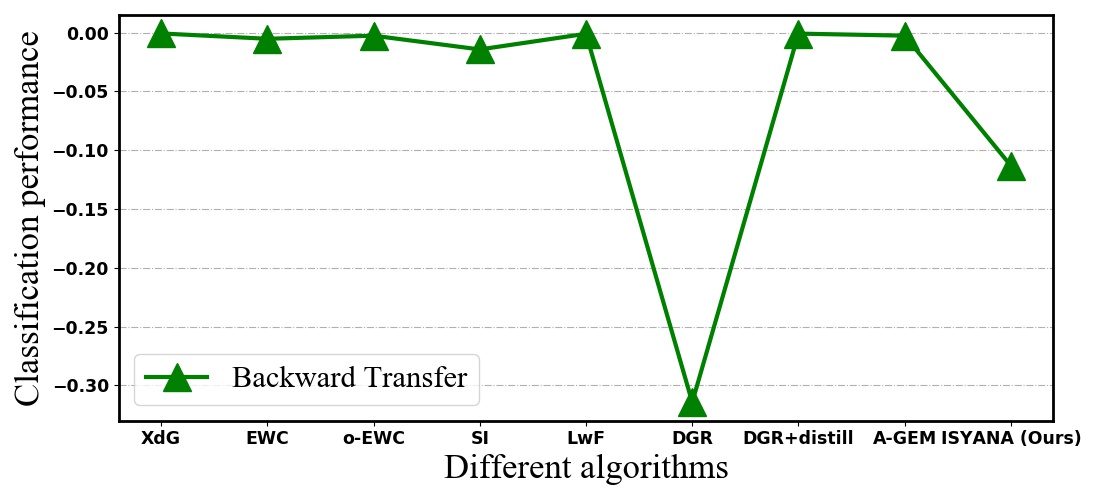}
}
\caption{Classification performance of different algorithms on splitMNIST.}
\label{fig:splitperformance}
\end{figure}

\begin{figure}[t]
\centering \subfigure[Average Accuracy.]{
   \label{fig:rotateacc}
   \includegraphics[width=1.\columnwidth]{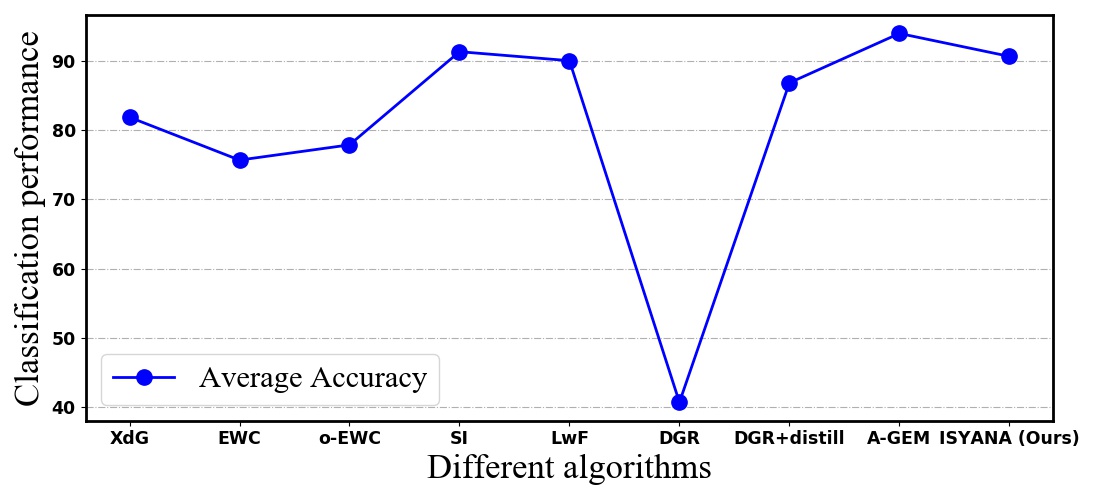}
} \subfigure[Forward Transfer.]{
   \label{fig:rotatefwt}
   \includegraphics[width=1.0\columnwidth]{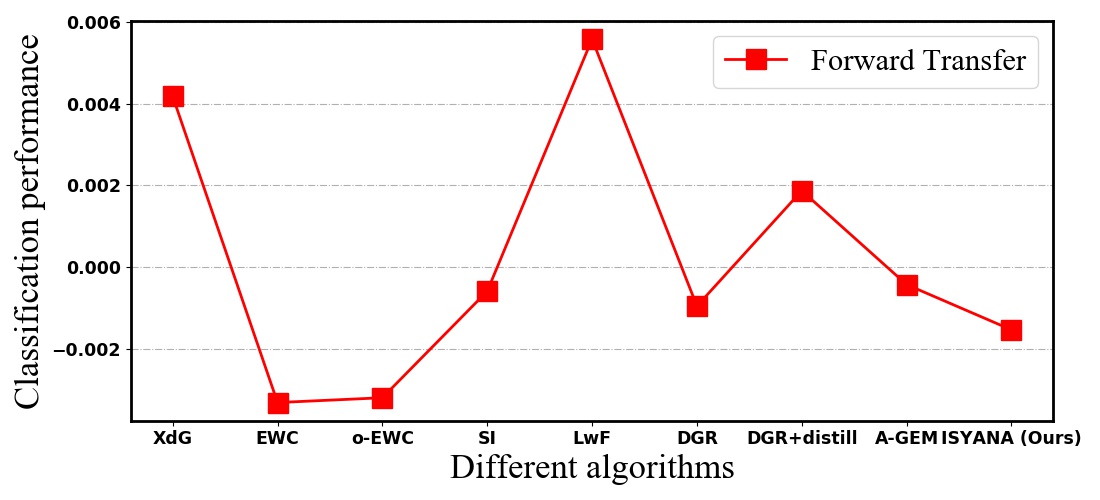}
} \subfigure[Backward Transfer.]{
   \label{fig:rotatebwt}
   \includegraphics[width=1.0\columnwidth]{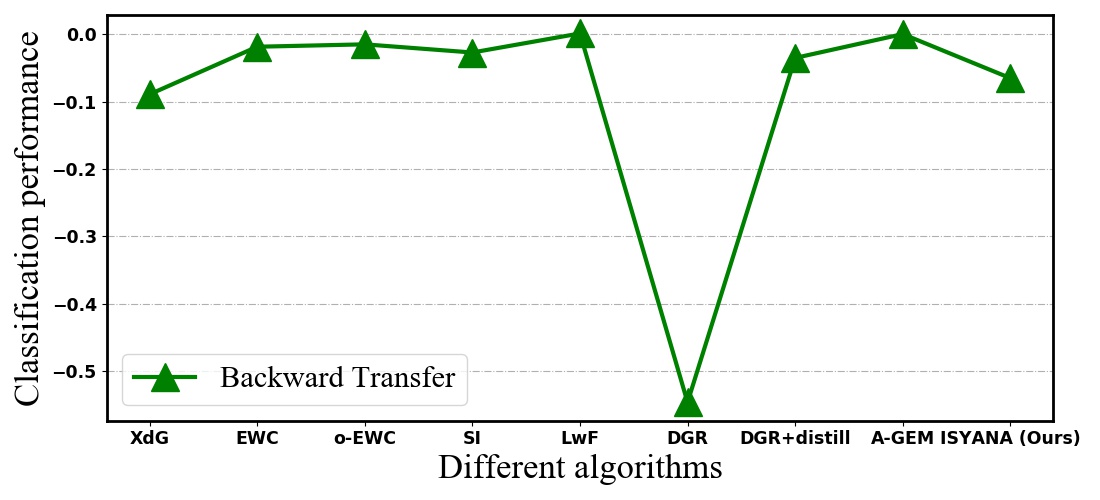}
}
\caption{Classification performance of different algorithms on rotatedMNIST.}
\label{fig:rotateperformance}
\end{figure}

\begin{figure}[t]
\centering \subfigure[Average Accuracy.]{
   \label{fig:omniglotacc}
   \includegraphics[width=1.\columnwidth]{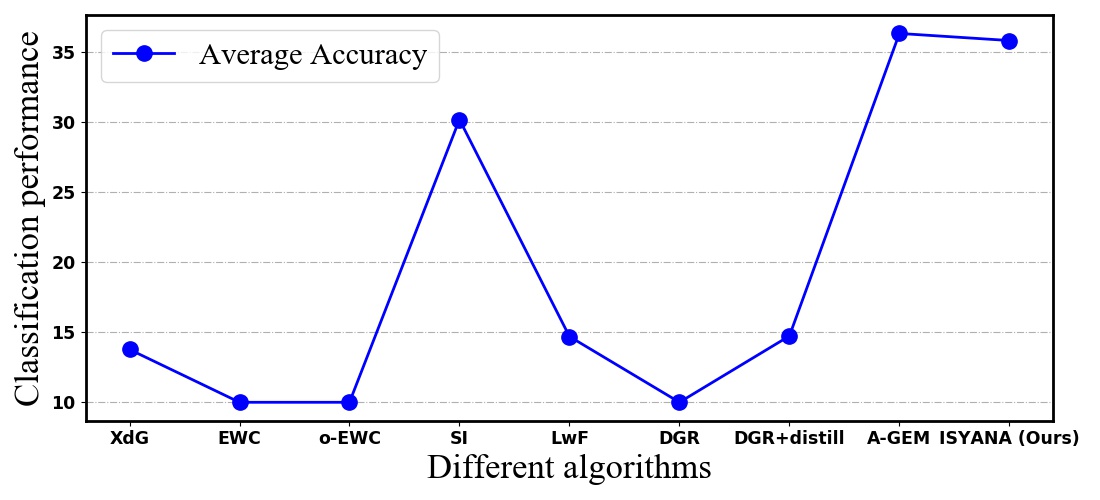}
} \subfigure[Forward Transfer.]{
   \label{fig:omniglotfwt}
   \includegraphics[width=1.0\columnwidth]{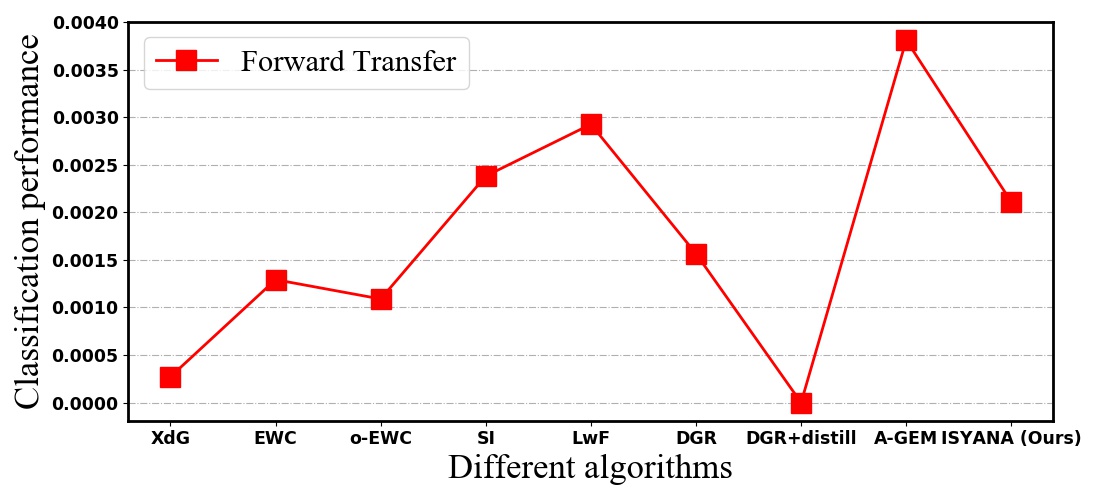}
} \subfigure[Backward Transfer.]{
   \label{fig:omniglotbwt}
   \includegraphics[width=1.0\columnwidth]{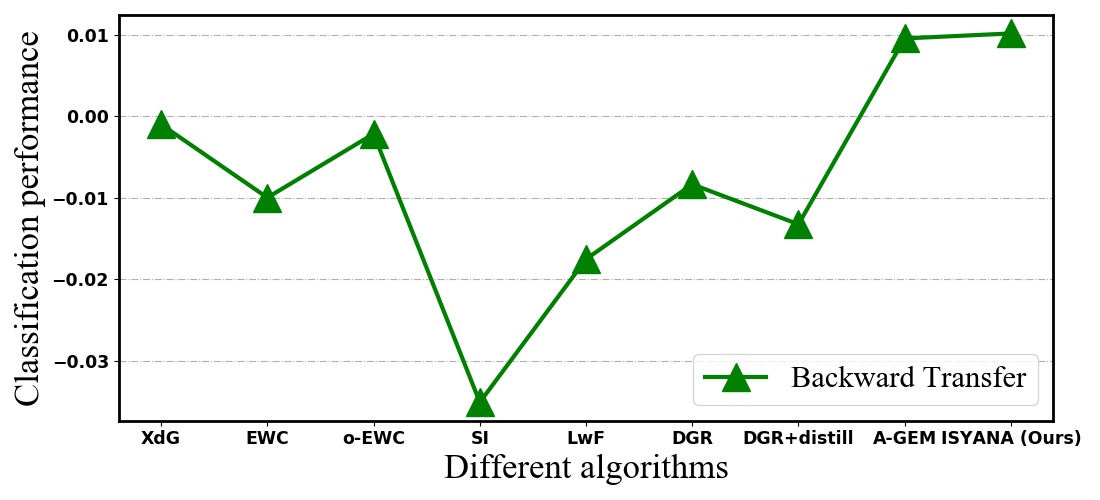}
}
\caption{Classification performance of different algorithms on omniglot.}
\label{fig:omniglotperformance}
\end{figure}

We implement our flow based on the work~\cite{vandeven2019three,vandeven2018generative} in the python programming language and perform our experiments on a Windows platform. We adopt the original 28x28 pixel grey-scale images without pre-processing~\cite{vandeven2019three,vandeven2018generative}. For SplitMNIST, the dataset is divided into five tasks where each task is a two-way classification. For permutedMNIST and rotatedMNIST, the dataset is split into ten tasks where each task is a ten-way classification respectively. For omniglot dataset, it consists of 50 tasks where each task encompasses 10 classes \cite{schwarz2018progress}. It is akin to the splitMNIST problem where each task is devoted to handle the incremental class problem. We run each simulation for 10 times with different random seeds and adopt the average value as the measurable criterion to evaluate the performance of each method. Table \ref{result-splitmnist}, \ref{result-permutedMNIST}, \ref{result-rotatedMNIST} and \ref{result-omniglot} corresponding to Fig.~\ref{fig:splitperformance}, Fig.~\ref{fig:permuteperformance}, Fig.~\ref{fig:rotateperformance} and Fig.~\ref{fig:omniglotperformance} show the numerical results, i.e., the ACC, BWT, and FWT among different approaches for splitMNIST, rotatedMNIST, permutedMNIST and omniglot respectively.

\section{Experimental Results}
\begin{figure}[t]
\centering \subfigure[Average Accuracy.]{
   \label{fig:permuteacc}
   \includegraphics[width=1.\columnwidth]{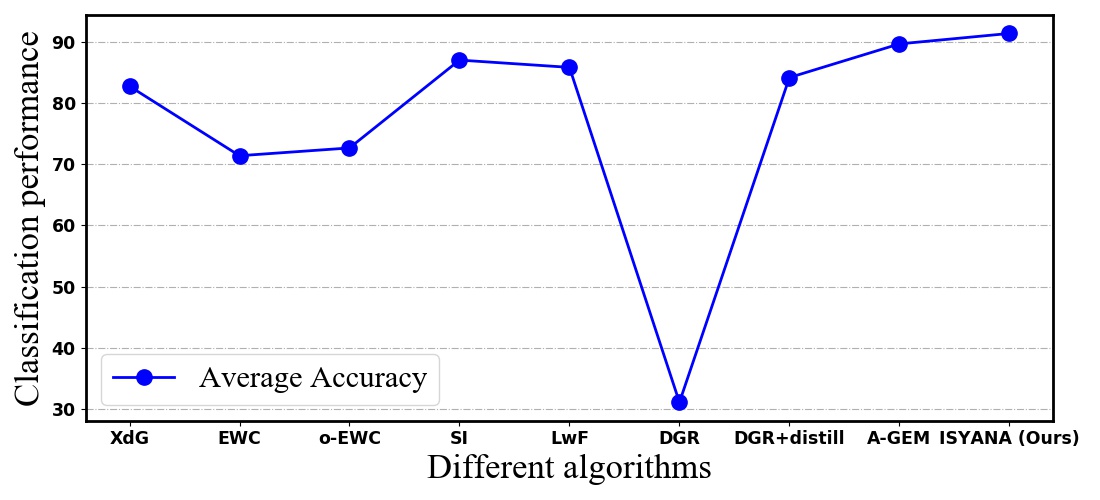}
} \subfigure[Forward Transfer.]{
   \label{fig:permutefwt}
   \includegraphics[width=1.0\columnwidth]{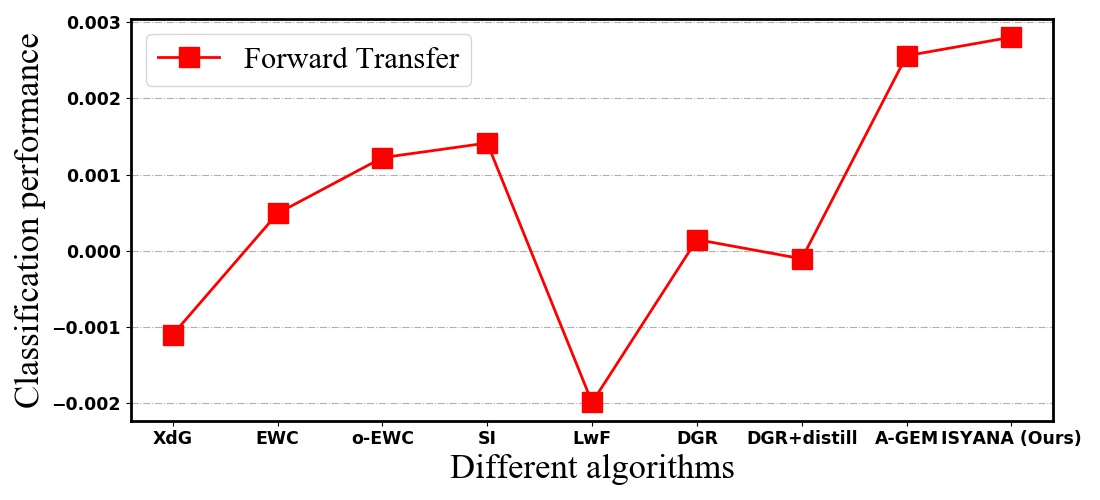}
} \subfigure[Backward Transfer.]{
   \label{fig:permutebwt}
   \includegraphics[width=1.0\columnwidth]{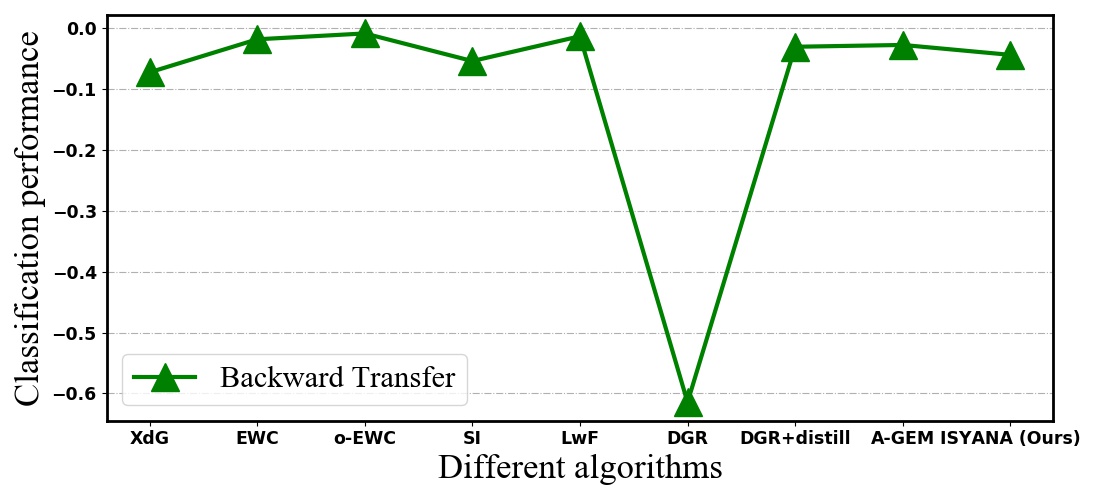}
}
\caption{Classification performance of different algorithms on permutedMNIST.}
\label{fig:permuteperformance}
\end{figure}

Table \ref{result-permutedMNIST} corresponding to Fig.~\ref{fig:permuteperformance} demonstrate that ISYANA offers significant performance improvement over other algorithms in the permuttedMNIST problem while achieving the highest FWT. This observation demonstrates the advantage of neuron-level plasticity control to cope with the changing data distributions. The use of synaptic-to-task mapping allows flexible activation and deactivation of a node in respect to their relevance to a task. On the other hand, the task-to-task mapping measuring the inter-task relationship is capable of supporting positive forward and backward transfer. That is, an important node of previous tasks still accepts the incoming task provided high commonalities between these tasks. On the other hand, ISYANA attains comparable numerical results to SI and A-GEM in the rotatedMNIST problem. Notwithstanding that A-GEM delivers the highest performance here, A-GEM exploits past samples for sample replay mechanism. The advantage of ISYANA for large-scale continual learning problem is demonstrated in the Omniglot problem where its performance outperforms other algorithms except A-GEM with noticeable margin. It should be interpreted carefully since A-GEM utilizes external memory for sample replay whereas ISYANA is memory-free. ISYANA also has better BWT than A-GEM in this context. This finding is supported by the fact that ISYANA supports the positive forward and backward transfers across the task. Moreover, it is perceived that the explosion $Z$ is apparent here for other methods hindering for its deployment in the large-scale setting. Note that the omniglot problem comprises 50 tasks and 500 classes in total. 

ISYANA is, however, inferior to A-GEM, EWC, o-EWC and SI in the splitMNIST problem. We argue that this is as a result of the class-dependent nature of ISYANA where incremental classes are presented here. This leads to inaccurate estimation of neuron importance and inter-task relationship. 


\begin{table}[!ht]
\caption{Classification performance on splitMNIST dataset}
\label{result-splitmnist}
\begin{center}
\scalebox{1.0}{
\begin{sc}
\begin{tabular}{lcccr}
\toprule
Model & Acc. (\%) & FWT & BWT \\
\midrule
XdG & 60.22 & 0.000842 & -0.000777 \\
EWC & 92.49 & 0.003571 & -0.005320  \\
o-EWC & 92.73 & 0.002913 & -0.002684 \\
SI & 92.32 & 0.003936 & -0.014379 \\
LwF & 83.54 & -0.001945 & -0.001065 \\
DGR & 50.75 & -0.003138 & -0.314305 \\
DGR+distill & 83.71 & -0.001902 & -0.000926 \\
\textbf{A-GEM} & 92.37 & -0.001215346 & -0.002667118 \\
\textbf{ISYANA (Ours)} & \textbf{89.48} & \textbf{-0.027} & \textbf{-0.11323} \\
\bottomrule
\end{tabular}
\end{sc}}
\end{center}
\end{table}

\begin{table}[!ht]
\caption{Classification performance on permutedMNIST dataset}
\label{result-permutedMNIST}
\begin{center}
\scalebox{1.0}{
\begin{sc}
\begin{tabular}{lcccr}
\toprule
Model & Acc. (\%) & FWT & BWT \\
\midrule
XdG & 82.70 & -0.001100 & -0.072413 \\
EWC & 71.36 & 0.000491 & -0.018104  \\
o-EWC & 72.63 & 0.001222 & -0.008469 \\
SI & 86.97 & 0.001414 & -0.053851 \\
LwF & 85.78 & -0.001990 & -0.013095 \\
DGR & 31.12 & 0.000147 & -0.614638 \\
DGR+distill & 84.12 & -0.000105 & -0.030367 \\
\textbf{A-GEM} & 89.61 & 0.002559 & -0.027380 \\
\textbf{ISYANA (Ours)} & \textbf{91.3175} & \textbf{0.0028} & \textbf{-0.0436} \\
\bottomrule
\end{tabular}
\end{sc}}
\end{center}
\end{table}

\begin{table}[!ht]
\caption{Classification performance on rotatedMNIST dataset}
\label{result-rotatedMNIST}
\begin{center}
\scalebox{1.0}{
\begin{sc}
\begin{tabular}{lcccr}
\toprule
Model & Acc. (\%) & FWT & BWT \\
\midrule
XdG & 81.88 & 0.004178 & -0.089339 \\
EWC & 75.69 & -0.003319 & -0.018699  \\
o-EWC & 77.89 & -0.003201 & -0.015153 \\
SI & 91.35 & -0.000592 & -0.027260 \\
LwF & 90.05 & 0.005582 & 0.001019 \\
DGR & 40.70 & -0.000957 & -0.546055 \\
DGR+distill & 86.85 & 0.001869 & -0.035341 \\
\textbf{A-GEM} & 93.99 & -0.000430 & 0.000082 \\
\textbf{ISYANA (Ours)} & \textbf{90.6706} & \textbf{-0.0015375} & \textbf{-0.065291667} \\
\bottomrule
\end{tabular}
\end{sc}}
\end{center}
\end{table}



\begin{table}[!ht]
\caption{Classification performance on omniglot dataset}
\label{result-omniglot}
\begin{center}
\scalebox{1.0}{
\begin{sc}
\begin{tabular}{lcccr}
\toprule
Model & Acc. (\%) & FWT & BWT \\
\midrule
XdG	&	13.774	&	0.000272	&	-0.00102	\\
EWC	&	10	&	0.001292	&	-0.01	\\
o-EWC	&	10	&	0.001089	&	-0.00218	\\
SI	&	30.208	&	0.002381	&	-0.0351	\\
LwF	&	14.7	&	0.002925	&	-0.01755	\\
DGR	&	10	&	0.001565	&	-0.00837	\\
DGR+distill	&	14.74	&	0	&	-0.01327	\\
\textbf{A-GEM} &36.36 &	0.00381 & 0.009524 \\
\textbf{ISYANA (Ours)} & \textbf{35.85333} & \textbf{0.002109} & \textbf{0.010136} \\
\bottomrule
\end{tabular}
\end{sc}}
\end{center}
\end{table}


\begin{table}[!ht]
\caption{Performance comparison on splitMNIST,permutedMNIST,rotatedMNIST and omniglot dataset}
\label{ourResultCompareWithNoTT}
\begin{center}
\scalebox{1.0}{
\begin{sc}
\begin{tabular}{lcccr}
\toprule
\multicolumn{4}{c}{splitMNIST} \\
\midrule
Model & Acc. (\%) & FWT & BWT \\
\midrule
ISYANA(No TT) & 89.062 & 0.012268 & -0.1177148 \\
\textbf{ISYANA (Ours)} & \textbf{89.48} & \textbf{-0.027} & \textbf{-0.11323} \\
\midrule
\multicolumn{4}{c}{permutedMNIST} \\
\midrule
Model & Acc. (\%) & FWT & BWT \\
\midrule
ISYANA(No TT) & 90.372 & 0.0002288 & -0.0520135 \\
\textbf{ISYANA (Ours)} & \textbf{91.3175} & \textbf{0.0028} & \textbf{-0.0436} \\
\midrule
\multicolumn{4}{c}{rotatedMNIST} \\
\midrule
Model & Acc. (\%) & FWT & BWT \\
\midrule
ISYANA(No TT) & 90.136 & -0.0040442 & -0.0674123 \\
\textbf{ISYANA (Ours)} & \textbf{90.6706} & \textbf{-0.0015375} & \textbf{-0.065291667} \\
\midrule
\multicolumn{4}{c}{omniglot} \\
\midrule
Model & Acc. (\%) & FWT & BWT \\
\midrule
ISYANA(No TT) & 9.99 & -0.0001021 & -0.0005102 \\
\textbf{ISYANA (Ours)} & \textbf{35.85333} & \textbf{0.002109} & \textbf{0.010136} \\
\bottomrule
\end{tabular}
\end{sc}}
\end{center}
\end{table}

\section{Ablation Study}
This section aims to study the effect of learning components to the overall learning performance of ISYANA. The underlying interest is to assess the inter-task nature of ISYANA where shareable information of each task is exploited to enable positive forward and backward transfer. That is, the tast-to-task mapping is deactivated such that the learning process is solely guided by the synaptic-to-task mapping. This study is carried out using all four datasets: splitMNIST, rotatesMNIST, permuttedMNIST and omniglot. Table \ref{ourResultCompareWithNoTT} reports our results.

From Table \ref{ourResultCompareWithNoTT}, the advantage of the task-to-task module is obvious in which its absence results in performance's deterioration. It reduces the classification rate of ISYANA by around $0.4\% - 0.5\%$ in the splitMNIST problem and the rotatedMNIST problem. On the other hand, there does not exist any performance difference in the splitMNIST problem. Dramatic performance degeneration exists in the omniglot problem where ISYANA's accuracy decreases by around $25\%$. This finding demonstrates the influence of the task-to-task component for a large-scale continual learning problem. It aligns with the fact that the task-to-task component is capable of exploiting the inter-task relationship which can be shared across important neurons.

\section{Other Approaches}
In the continual learning domain, there exists other approaches to combat the catastrophic forgetting approach encompassing the memory-based approach and the structure-based approach. It is outlined as follows:
\newline\noindent\textbf{Memory-based Approach}: the catastrophic forgetting problem is solved here by utilizing external memory of past samples to be replayed such that the catastrophic forgetting problem can be overcome. iCaRL is a representative approach utilizing the external memory in addressing the catastrophic forgetting problem \cite{Rebuffi2017iCaRLIC}. It creates exemplar sets representing each class and the classification decision is performed by checking the similarity of a sample to the most similar exemplar set. Gradient Episodic Memory \cite{gem2017paz} and its extension A-GEM \cite{Chaudy2019AGEM} are categorized as the memory-based approaches. GEM utilizes an episodic memory \cite{gem2017paz} storing a subset of the observed samples where the forgetting case is indicated by measuring the angle of the gradient vector and the proposed update. A-GEM modifies the loss function of GEM expediting the model update. Deep Generative Replay (DGR) \cite{shin2017continual} does not store previous samples to alleviate the catastrophic forgetting problem rather utilizes generative adversarial network (GAN) to generate pseudo-data preventing the catastrophic forgetting problem. This work is extended in \cite{vandeven2019three} with the knowledge distillation and feedback connection. Since the memory-based approach utilizes the experience-replay mechanism to address the catastrophic forgetting problem, it is capable of handling the single-head continual learning scenario where other approaches cannot deliver. Nevertheless, compared to regularization-based approach, the memory-based approach is computationally prohibitive. Furthermore, choosing which samples to retain remains an open question because of the dynamic learning environments. 
\newline\noindent\textbf{Structure-based Approach}: another approach in the continual learning uses the structure-based approach where it freezes some components of a model while introducing a new component to embrace new task. Progressive Neural Network \cite{Rusu2016ProgressiveNN} is a pioneering work in this domain where a new column is inserted for new task while freezing the old components to avoid the catastrophic forgetting problem. Dynamically Expandable Networks (DEN) \cite{Lee2018LifelongLW} is another technique in this domain where it addresses some flaws of PNN. It adopts the selective retraining approach and the splitting/duplicating strategy. Recently, the so-called learn-to-grow framework is put forward in \cite{learntogrow}. It adopts the neural architecture search to find a network structure that best handles each task. The underlying limitation of the structure-based approach for continual learning is seen in its high computational and memory burdens.

\section{Conclusion}
An inter-task synaptic mapping (ISYANA) is proposed here to perform the continual learning of data streams having the dynamic and evolving characteristics. ISYANA is underpinned by two components, the task-to-task and synaptic-to-task mapping components where the task-to-task mapping is developed to study the mutual information across tasks while the synaptic-to-task mapping focuses on the mapping between a node to a task. This trait enables exploitation of shareable information across task while retaining private information thereby enhancing the forward and backward transfers. The advantage of ISYANA has been numerically validated in the benchmark problems and recently published algorithms clearly showing improved performance. Our future work targets an extension of ISYANA for purely streaming environments where there exist uncertainties for task's boundary.

\section*{Acknowledgment}
This research is financially supported by National Research Foundation, Republic of Singapore under IAFPP in the AME domain (contract no.: A19C1A0018). This work was mainly done when the first author was a research fellow in NTU.

\bibliography{mybibfile}

\appendix
\section{Experimental Details}
The detail of each experiments are shown in Table \ref{hyper-para mnist} and Table \ref{Hyper-omniglot}. For EWC, Online-EWC and SI, the loss function includes a regularization term, the loss value controlled by a strength parameter: $ {L}_{total}=L_{current}+ \lambda \times {L}_{reg}$. For sample replayed methods, LwF, DGR and DGR+distill, the loss for replayed data is added the the loss function. 
In Omniglot experiment, there are 50 alphabets in Omniglot dataset, the task is divided based on the alphabets. The dataset is augmented by 20 kinds of transformations, 10 rotations and 10 shifting. In our experiments, 10 classes was chosen in each task. The result of fisher matrix calculation for EWC and Onine EWC increases dramatically along with the training steps, which causes the poor results.

\begin{table}
\caption{Hyper-parameters for the experiments on split MNIST, rotate MNIST and permutate MNIST.}
\label{hyper-para mnist}
\begin{center}
\scalebox{1.0}{
\begin{sc}
\begin{tabular}{lcccr}
\toprule
Experiment & Split MNIST & Rotate MNIST & Permuted MNIST\\ 
\midrule
Network(MLP) & {[784,256,256]} & {[784,500,500]} & {[784,500,500]} \\
Optimizer & \multicolumn{3}{c}{SGD learning rate:0.1}\\
Batch size &  \multicolumn{3}{c}{128} \\
Training epochs &  \multicolumn{3}{c}{10}\\
\midrule
XdG &  \multicolumn{3}{c}{80 percentage neurons per layer to gate} \\
EWC &  \multicolumn{3}{c}{lambda:5000}\\
o-EWC &  \multicolumn{3}{c}{lambda:5000; forgetting coefficient:1.0}\\
SI &  \multicolumn{3}{c}{regularisation strength:1 Dampening ratio:0.1}\\
DGR and DGR+distill &  \multicolumn{3}{c}{VAE:MLP[784,500,500]}\\
A-GEM &  \multicolumn{3}{c}{replay samples from previous task:2000}\\
ISYANA(ours) & \multicolumn{3}{c}{$\eta = 1 *\exp{(-Z-0.1)}$} \\
\bottomrule
\end{tabular}
\end{sc}}
\end{center}
\end{table}

\begin{table}
\caption{Hyper-parameters for Omniglot Experiment}
\label{Hyper-omniglot}
\begin{center}
\scalebox{1.0}{
\begin{sc}
\begin{tabular}{lcccr}
\toprule
Parameter\\ 
\midrule
Network(MLP) & {[784,500,500]} \\
Optimizer & SGD learning rate:0.1 (except ISYANA)\\
Batch size & 32 \\
Training epochs & 5 \\
\midrule
XdG & 80 percentage neurons per layer to gate \\
EWC & lambda:5000\\
o-EWC & lambda:5000\\
SI & regularisation strength:0.1 Dampening ratio:0.1\\
DGR and DGR+distill & VAE:MLP[784,500,500]\\
A-GEM & replay samples from previous task:2000\\
ISYANA(ours) & $\eta = 1 *\exp{(-Z-0.1)}$ \\
\bottomrule
\end{tabular}
\end{sc}}
\end{center}
\end{table}

\end{document}